\documentclass[journal,transmag]{IEEEtran}
\usepackage{booktabs}
\usepackage{multirow}

\usepackage{cite}

\usepackage{graphicx}
\ifCLASSINFOpdf
\else
\fi
\usepackage{amsmath}
\usepackage{amssymb}
\usepackage{caption}
\DeclareMathOperator*{\argmax}{arg\,max}

\begin{document}
%
\title{Leaving Some Stones Unturned:\\Dynamic Feature Prioritization\\for Activity Detection in Streaming Video}
%
%
%

\author{
\IEEEauthorblockN{Yu-Chuan~Su and
Kristen~Grauman}
\IEEEauthorblockA{Department of Computer Science, The University of Texas at Austin}
}

\maketitle

\begin{abstract}
Current approaches for activity recognition often ignore constraints on computational resources: 1) they rely on extensive feature computation to obtain rich descriptors on all frames, and 2) they assume batch-mode access to the entire test video at once.  We propose a new \emph{active} approach to activity recognition that prioritizes ``what to compute when" in order to make timely predictions.  The main idea is to learn a policy that dynamically schedules the sequence of features to compute on selected frames of a given test video.
In contrast to traditional static feature selection, our approach continually re-prioritizes computation based on the accumulated history of observations and accounts for the transience of those observations in ongoing video.  We develop variants to handle both the batch and streaming settings.  On two challenging datasets, our method provides significantly better accuracy than alternative techniques for a wide range of computational budgets.
\end{abstract}


\section{Introduction}

Activity recognition in video is a core vision challenge.  It has applications
in surveillance, autonomous driving, human-robot interaction, and automatic
tagging for large-scale video retrieval.  In any such setting, a system that
can both categorize and temporally localize activities would be of great value.

Activity recognition has attracted a steady stream of interesting
research~\cite{ryoo-survey}.  Recent methods are largely learning-based, and
tackle realistic everyday activities (e.g., making tea, riding a bike).  Due to
the complexity of the problem, as well as the density of raw data comprising
even short videos, useful video representations are often computationally
intensive---whether dense trajectories, interest points, object detectors,  or
convolutional neural network (CNN) features run on each
frame~\cite{wang2013iccv,pirsiavash2012cvpr,jain2015cvpr,simonyan-activity,hauptman-cvpr2015,zha-bmvc2015,han2009iccv}.
In fact, the expectation is that \emph{the more features one extracts from the
video, the better for accuracy}.  For a practitioner wanting reliable activity
recognition, then, the message is to ``leave no stone unturned", ideally
extracting complementary descriptors from all video frames.

However, the ``no stone unturned" strategy is problematic.  Not only does it
assume virtually unbounded computational resources, it also assumes that an
entire video is available at once for batch processing.  In reality, a
recognition system will have some computational budget.  Further, it may need
to perform in a \emph{streaming} manner, with access to only a short buffer of
recent frames.   Together, these considerations suggest some form of feature
triage is needed.

Yet prioritizing features for activity in video is challenging, for two key
reasons.  First, the most informative features may depend critically on what
has been observed so far in the specific test video, making traditional
fixed/static feature selection methods inadequate.  In other words, the
recognition system's belief state must evolve over time, and its priorities of
which features to extract next must evolve too.  Second, when processing
streaming video, the entire video is never available to the algorithm at once.
This puts limits on what features can even be considered each time step, and
requires accounting for the feature extractors' framerates when allocating
computation.

In light of these challenges, we propose a dynamic approach to prioritize
\emph{which features to compute when} for activity recognition.  We formulate
the problem as policy learning in a Markov decision process.  In particular, we
learn a non-myopic policy that maps the accumulated feature history (state) to
the subsequent feature and space-time location (action) that, once extracted,
is most expected to improve recognition accuracy (reward) over a sequence of
such actions.  We develop two variants of our approach: one for batch
processing, where we are free to ``jump" around the video to get the next
desired feature, and one for streaming video, where we are confined to a buffer
of newly received frames.  By dynamically allocating feature extraction effort,
our method wisely leaves some stones \emph{unturned}---that is, some features
unextracted---in order to meet real computational budget constraints.

To our knowledge, our work is the first to actively triage feature computation
for streaming activity recognition.\footnote{This paper extends our earlier
technical report~\cite{AI15-05}.}  While recent work explores ways to
intelligently order feature computation in a static image for the sake of
object or scene
recognition~\cite{butko,sudheendra-cvpr2010,karayev-nips,dulac2014iclr,karayev2014cvpr,gonzalez2015cvpr,gao2011nips,yu2011cvpr}
or offline batch activity detection~\cite{yeung2015end}, streaming video
presents unique challenges, as we explain in detail below.  While methods for
``early" detection can fire on an action prior to its
completion~\cite{hoai2012cvpr,ryoo2011iccv,davis-tyagi-2006}, they nonetheless
passively extract all features in each incoming frame.

We validate our approach on two public datasets consisting of third- and
first-person video from over 120 activity categories.  We show its impact in
both the streaming and batch settings, and we further consider scenarios where
the test video is ``untrimmed".  Comparisons with status quo passive feature
extraction, traditional feature selection approaches, and a state-of-the-art
early event detector demonstrate the clear advantages of our approach.

\section{Related Work}

\subsection{Activity recognition and detection}

Recognizing activities is a long-standing vision challenge~\cite{ryoo-survey}.
Current methods explore both high-level representations based on objects,
attributes, or
scenes~\cite{han2009iccv,pirsiavash2012cvpr,feifei-yao-iccv2011,rohrbach-eccv2012,jain2015cvpr},
as well as holistic frame-level CNN
descriptors~\cite{jain2015cvpr,simonyan-activity,hauptman-cvpr2015,zha-bmvc2015}.
Our approach is a general algorithm for feature prioritization, and it is
flexible to the descriptor type; we demonstrate instances of both types in our
results.  Unlike traditional activity recognition work, we account for 1)
bounded computational resources for feature extraction and 2) streaming (and
possibly untrimmed) input video.

Much less work addresses activity \emph{detection}, which requires both
categorizing and localizing an activity in untrimmed video. Common strategies
are sliding temporal window
search~\cite{ye-hebert-iccv2005,duchenne-iccv2009,satikin-eccv2010}  or
analyzing tracked
objects~\cite{medioni-stream,yao-cvpr2010,klaser-human-focused-2010,mori-iccv2011}.
While some tracking-based methods permit incremental computation and thus can
handle streaming video (e.g.,~\cite{medioni-stream}), they are limited to
activities well-defined by a moving foreground subject.  ``Action-like"
space-time
proposals~\cite{yu-yuan-fast-cvpr2015,jain-cvpr2015,gkioxari2015cvpr,gemert2015bmvc}
and efficient search methods~\cite{chaoyeh-cvpr2012,yu-subvolume-cvpr2011} can
avoid applying classifiers to all possible video subvolumes, but they do not
prioritize feature computation.  A recurrent neural network learns to predict
which frame in a video to analyze next for offline action
detection~\cite{yeung2015end}; its policy is free to hop forward and backward
in time in the video to extract subsequent features, which is not possible in
the streaming case we consider.  Furthermore, our method pinpoints feature
extraction requests to include not just when in the video to look for a single
type of feature~\cite{yeung2015end}, but also where in the frame to look and
which particular feature to extract upon looking there.  Unlike our approach,
all the above prior classifier-based methods assume batch access to the entire
test video.  Furthermore, with the exception of~\cite{yeung2015end}, they also
assume features can be extracted on every frame.

\subsection{Early event detection}

The goal in ``early" event detection is for the detector to fire early on in
the activity instance, enabling timely reactions (e.g., for human-robot
interactions~\cite{hoai2012cvpr} or nefarious activity in
surveillance~\cite{ryoo2011iccv}).  In~\cite{hoai2012cvpr}, a structured output
approach learns to recognize partial events in untrimmed video.  Other methods
tackle trimmed streaming video, developing novel integral-histograms that
permit incremental recognition~\cite{ryoo2011iccv}, or an HMM model that
processes more frames until its action prediction is
trusted~\cite{davis-tyagi-2006}.  In a sense, ``early" detectors eliminate
needless computation.  However, the goals and methods are quite different from
ours.  They intend to detect an action before its completion, whereas we aim to
detect an action with limited computation.  As such, whereas the early methods
``front-load" computation---extracting all features for each incoming
frame---our method targets \emph{which features to compute when}, and can even
skip frames altogether.  Furthermore, rather than learn a static model of what
the onset of an action looks like, we learn a dynamic policy that indicates
which computation to perform given past observations.

\subsection{Fast object detection}

Various ways to accelerate object detection have been
explored~\cite{sadeghi2014eccv,yan2014cvpr,viola-jones,cascade-vedaldi}.
Cascaded and coarse-to-fine detectors
(e.g.,~\cite{viola-jones,cascade-vedaldi}) determine a fixed ordering of
features to quickly reject unlikely regions.  In contrast, our work deals with
activity recognition in video, and the feature ordering we learn is dynamic,
non-myopic, and generalizes to streaming data.

\subsection{Active object and scene recognition in images}

Recent work considers ``active" and ``anytime" object recognition in
images~\cite{karayev2014cvpr,gonzalez2015cvpr,karayev-nips,gao2011nips,dulac2014iclr,butko,sudheendra-cvpr2010,yu2011cvpr,weiss2013nips}.
The goal is to determine which feature or classifier to apply next so as to
reduce inference costs and/or supply an increasingly confident estimate as time
progresses.  Several methods explore dynamic feature selection algorithms for
object and scene
recognition~\cite{karayev2014cvpr,gao2011nips,dulac2014iclr,yu2011cvpr}, using
strategies based on reinforcement
learning~\cite{karayev2014cvpr,karayev-nips,dulac2014iclr,weiss2013nips}, or
myopic information gain~\cite{yu2011cvpr,gao2011nips}.  Though focused on scene
recognition in images, \cite{yu2011cvpr} also includes a preliminary trial for
``dynamic scenes" in short trimmed videos; however, the model does not
represent temporal dynamics, the data is batch-processed, and gains over
passive recognition are not shown.   These existing methods categorize an image
(recognition), search for an object
(detection)~\cite{butko,gonzalez2015cvpr,karayev-nips,sudheendra-cvpr2010} or
perform structured prediction~\cite{weiss2013nips}.

This family of methods is most relevant to our goal.  However, whereas prior
work performs object/scene recognition in images, we consider activity
recognition in streaming video.  Feature triage on video offers unique
challenges.  Active recognition on images is a feature ordering task: one has
the entire image in hand for processing, and the results of selected
observations are static and simply accumulate.  In contrast, for video,
features come and go, and we must update beliefs over time and prioritize
future observations accordingly.  Furthermore, we must represent temporal
continuity (i.e., model context over both time and space) and, when streaming,
respect the hard limits of the video buffer size.  In terms of a Markov
decision process, this translates into a much larger state-action space.

\subsection{Allocating computation for video}

To our knowledge, no prior work studies dynamically prioritizing features for
streaming activity recognition, while there is limited work prioritizing
computation for other tasks in video.  In~\cite{karasev2014cvpr}, information
gain is used to determine which object detectors to deploy on which frames for
semantic segmentation.  In~\cite{chen2011pami}, a second-order Markov model
selects frames to apply a more expensive algorithm, for face detection and
background subtraction.  A cost-sensitive approach to multiscale video parsing
schedules inference at different levels of a hierarchy  (e.g., a group activity
composed of individual actions) using AND-OR
graphs~\cite{amer2012eccv,amer2013iccv}.  Aside from being different tasks than
ours, all the above methods consider only the offline/batch scenario.

\section{Approach}

We first formalize the problem (Sec.~\ref{sub:problem_formulation}).  Then we
present our approach and explain the details of its batch and streaming
variants (Sec.~\ref{sec:rl}).

\subsection{Problem Formulation}
\label{sub:problem_formulation}

Let $X \in \mathcal{X}$ denote a video clip and let $y \in \mathcal{Y}$ denote
an activity category label.  During training we have access to a set
$\{(X_1,y_1),\dots,(X_T,y_T)\}$ of video clips, each labeled by one of $L$
activity categories, $y_i \in \{1,\dots,L\}$.  The training clips are
temporally trimmed to the action of interest.  At test time, we are given a
novel video that may be trimmed or untrimmed.  For the trimmed case, the
ultimate goal is to predict the activity category label (i.e., a multi-way
recognition task).  For the untrimmed case, the goal is to temporally localize
when an activity appears within it (i.e., a binary detection
task).\footnote{For clarity of presentation, in the following we present our
method assuming a trimmed input video; Sec.~\ref{sec:streaming} explains
adjustments for untrimmed inputs.}

First, we train an activity recognition module using the labeled videos.  Let
$\Psi(X)$ denote a descriptor computed for video $X$.  We train an activity
classifier $f : \Psi \times \mathcal{Y} \rightarrow \mathbb{R}$ to return a
posterior for the specified activity category:
\begin{equation}
    f(\Psi(X),y) = P(y | X).\label{eq:classifier}
\end{equation}
We use one-vs-all multi-class logistic regression classifiers for $f$ and
bag-of-object or CNN descriptors for $\Psi$ (details below), though other
choices are possible.  When training $f$, descriptors on training videos are
fully instantiated using all frames.  This classifier is trained and fixed
prior to policy learning.

We formulate dynamic feature prioritization as a reinforcement learning
problem: the system must learn a policy to request the features in sequence
that will, over the course of a recognition \emph{episode}, maximize its
confidence in the true activity category.  At test time, given an unlabeled
video, inference is a sequential process.  At each step $k=1,\dots,K$ of an
episode we must 1) actively prioritize the next feature computation
\emph{action} and 2) refine the activity category prediction.  Thus, our
primary goal is to learn a \emph{dynamic policy} $\pi$ that maps partially
observed video features to the next most valuable action.  This policy should
be far-sighted, such that its choices account for interactions between the
current request and subsequent features to be selected.  Furthermore, it should
respect a \emph{computational budget}, meaning it conforms to constraints on
the feature request costs and/or the number of inference steps permitted.  We
consider both \emph{batch} and \emph{streaming} recognition settings.

\subsection{Learning the Feature Prioritization Policy}\label{sec:rl}

We develop a solution using a Markov decision process (MDP), which is defined by the following components~\cite{russell-norvig}:
\begin{itemize}
    \item A \textbf{state} $s_k$ that captures the current environment
         at the $k$-th step of the episode, defined in terms of the history of extracted features and prior actions.
    \item A set of discrete \textbf{actions} $\mathcal{A}=\{a_{m}\}^{M}_{m=1}$ the system can perform at each step in the
        episode, which will lead to an update of the state.   An action extracts information from the video.

    \item An instant \textbf{reward} $r_k = R(s_k, a^{(k)}, s_{k+1})$ received
        by transitioning from state $s_k$ to state $s_{k+1}$ after taking
        action $a^{(k)}$, defined in terms of activity recognition.  The total
        reward is $\sum_{k} \gamma^k R(s_k, a^{(k)}, s_{k+1})$, where $\gamma
        \in [0,1]$ is a discount factor on future rewards. Larger values lead
        to more far-sighted policies.

    \item A \textbf{policy} $\pi : s \rightarrow a$ determines the next action based on the current
        state. It selects the action that maximizes the expected reward:
        \begin{equation}
        \pi(s_k) = \argmax_{a} E[R|s_k,a,\pi],\label{eq:expected}
        \end{equation}
for this action and future actions continuing under the same policy.
 \end{itemize}

We next detail the video representation, state-action features, and rewards for
the general case.  Then, we define aspects specific to the batch and streaming
settings, respectively.

\paragraph{Video Descriptors and Actions}

Our algorithm accommodates a range of descriptor/classifier choices.  The
requirements are that the descriptor 1) have temporal locality, and 2) permit
incremental updates as new descriptor instances are observed.  These specs are
met by popular ``bag-of-X'' and CNN frame features, as we will demonstrate in
results, as well as others like quantized dense trajectories or human body
poses.

We focus our implementation primarily on a \emph{bag-of-objects} descriptor.
Suppose we have object detectors for $N$ object categories.  The fully observed
descriptor $\Psi(X)$ is an $N$-dimensional vector, where $\Psi_n(X)$ is the
likelihood that the $n$-th object appears (at least once) in the video clip
$X$.   We chose a bag-of-objects for its strength in compactly summarizing
high-level content relevant to
activities~\cite{jain2015cvpr,pirsiavash2012cvpr,gupta2007cvpr}.  For example,
an activity like ``making sandwich" is definable by bread, knife, frig, etc.
Furthermore, it exposes semantic temporal context valuable for sequential
feature selection.  For example, after seeing a mug, the system may learn to
look next for either a tea bag or a coffee maker.

Each step in an episode performs some action $a^{(k)}\in\mathcal{A}$ at a
designated time $t^{k}$ in the video.  We define each action as a tuple
$a_{m}=\langle o_{m},l_{m} \rangle$ consisting of an object and video
location.\footnote{Note that $a^{(k)}$ identifies an action selected at step
$k$ in the episode, whereas $a_{m}$ is one of the $M$ discrete action choices
in $\mathcal{A}$.}  Specifically, $o_m \in \{1,\ldots,N\}$ specifies an object
detector, and $l_m$ specifies the space-time subvolume where to run it.  The
observation result $x_{m}$ of taking action $a_{m}$ is the maximum detection
probability of object $o_{m}$ in volume $l_{m}$.\footnote{Some object detectors
share features across object categories, e.g., R-CNN~\cite{girshick2014cvpr}, 
in which case it may be practical to simplify the action to select only the 
video volume and apply all object classes. We use the DPM
detector~\cite{felzenszwalb2010pami}, which has the advantage of near real-time
detection~\cite{yan2014cvpr} using a single thread, whereas R-CNN relies
heavily on parallel computation and hardware acceleration~\cite{ren2015nips}.}
It is used to incrementally refine the video representation $\Psi(X)$.  Let
$o^{(k)}=n$ denote the object specified by selected action $a^{(k)}$.  Upon
receiving $x^{(k)}$, the $n$-th entry in $\Psi(X^{k})\in\mathbb{R}^{N}$ is
updated by taking the maximum observed probability for that object so far:
\begin{equation}
\Psi_n(X^k) = \max\left(\Psi_n(X^k), x^{(k)}\right),
    \label{eq:bag-of-object}
\end{equation}
where $\Psi(X^k)$ denotes the video representation based on the observation
results up to the $k$-th step of the episode. The initialization of $\Psi(X)$
is explained below.

To alternatively apply our method with CNN features---which show promise for
video (e.g.,~\cite{hauptman-cvpr2015,zha-bmvc2015,simonyan-activity})---we
define the representation and actions as follows.  The video representation
averages per-frame CNN descriptors:
\begin{equation}
    \Psi_n(X^k) = \text{mean}\left(X^{k}\right),
\end{equation}
and the action becomes $a_m = l_m$, since we need to specify the temporal
location alone.  Though very fast CNN extraction is possible (76 fps on a
CPU~\cite{nvidia}), conventional approaches still require time linear in the
length of the video, since they touch each frame.  We offer \emph{sub-linear}
time extraction; for example, our results maintain accuracy for streaming
recognition with CNNs while pulling the features from fewer than 1\% of the
frames.

\paragraph{State-Action Features}

With Q-learning~\cite{russell-norvig}, the value of actions $E[R | s, a, \pi]$
in Eq.~\eqref{eq:expected} is evaluated with $Q^{\pi}(s,a)$.  It must return a
value for any possible state-action pair.  Our state space is very
large---equal to the number of possible features times the number of possible
space-time locations times their possible output values.  This makes exact
computation of $Q^{\pi}(s,a)$ infeasible.  Thus, as common in such complex
scenarios, we adopt a linear function approximation $Q^{\pi}(s,a) =
\theta^{T}\phi(s,a)$, where $\phi(s,a)$ is a feature representation of a
\emph{state-action} pair and $\theta$ is learned from activity-labeled training
clips (explained below).

The state-action feature $\phi(s,a)$ encodes information relevant to policy
learning: the \emph{previous object detection} results and the \emph{action
history}.  Past object detections help the policy learn to exploit object
co-occurrences (e.g., that running a laptop detector after finding soap is
likely wasteful) and select discriminative but yet-unseen objects (e.g., having
seen a chair, looking next for a bed or dish could disambiguate the bedroom or
kitchen context, whereas a cell phone would not).  The action history can also
benefit the policy, letting it learn to avoid redundant selections.

Motivated by these requirements, we define the state-action feature
$\phi(s,a) \in \mathbb{R}^{N+M}$ as
\begin{equation}
    \phi(s_k,a) = [\Psi(X^{k}), \delta t^{k}],
\end{equation}
where $\Psi(X^{k})$ encodes the detection results and $\delta t^{k}$
encodes the action history.
$\Psi(X^{k}) \in \mathbb{R}^{N}$ is the representation defined above.
The action history feature $\delta t^{k} \in
\mathbb{R}^{M}$ encodes how long it has been since each action was performed in the episode,
which for action $m$ is
\begin{equation}
    \delta t^{k}(m) = t^{k} - \max_{i}\{t^{i}|a^{(i)}=a_{m}\},
\end{equation}
with  $\delta t^{k}(m)=0$ if $a_{m}$ has never been performed before.

To encode actions into the state-action representation $\phi(s,a)$, we
learn one linear model $\theta_{a_{m}}$ for each action (details below), such
that $Q^{\pi}(s,a_{m})=\theta^{T}_{a_{m}}\phi(s,a)$. In the following, we denote
$\theta{=}\{\theta_{a_m}\}^{M}_{m=1}$.

\paragraph{Reward}

We define a smooth reward function that rewards increasing confidence in the
correct activity label, our ultimate prediction task.  Intuitively, the model
should continuously gather evidence for the activity during the episode, and
its confidence in the correct label should increase over time and surpass all
other activities by the time the computation budget is exhausted.  Accordingly,
for a training episode run on video $X$ with label $y^\ast$, we define the
reward:
\begin{equation}
    R(s_k, a^{(k)}, s_{k+1}) = f(\Psi(X^{k+1}),y^\ast) - f(\Psi(X^{k}),y^\ast).\label{eq:reward}
\end{equation}
With this definition, a new action gets no ``credit" for confidence
attributable to previous actions. We found that rewarding accuracy increases
per unit time performs similarly to training multiple policies targeting fixed
budgets.  Moreover, the proposed reward has the advantage that we can run the
policy for as long as desired at test time, which is essential for streaming
video.  Fixed-budget policies, though common in RL, are ill-suited for
streaming data since we cannot know in advance the test video's duration and
the budget to allocate.

\paragraph{Dynamic Feature Prioritization Policy}

We learn the policy $\pi$ using policy iteration~\cite{russell-norvig}.
Policy iteration is an iterative algorithm that alternates between generating
training samples given a policy $\pi^{(i)}$ parametrized by $\theta^{(i)}$ and
learning $\theta^{(i+1)}$ given the generated training samples. We describe the
steps within one iteration next.

Given the policy $\pi^{(i)}$ learned from the previous iteration, new training
samples are generated by running recognition episode on all videos following
$\pi^{(i)}$. For each video, the recognition episode will result in a series of
three tuple
$\{(a^{(k)},\phi(s_{k},a^{(k)}),r_{k})\}_{k=1}^{K_j}$, where
the length $K_j$ is the number of actions performed when recognizing video $v_j$.
Each three tuple corresponds to one action in the episode, and we collect the
corresponding action, state-action-feature and reward during recognition. The
target value for $Q^{\pi}(s,a)$ can be computed as
\begin{equation}
    E[R|s_{k},a,\pi] = \sum_{k}^{K_j} \gamma^{k}r_{k},
\end{equation}
following the definition of total reward
after finishing the recognition episode.
Therefore, we can transform the three tuples into
$(a_{k},\phi(s_{k},a^{(k)}),E[R|s_k,a^{(k)},\pi])$, and learning $\theta^{(i+1)}$
from the three tuples becomes a regression problem 
\begin{equation}
    E[R|s_{k},a^{(k)},\pi] = \theta^{T}_{a^{(k)}}\phi(s_{k},a^{(k)}),
\end{equation}
where we solve it using ridge regression. The algorithm then
iterates, generating new samples using $\theta^{(i+1)}$. We
run a fixed number of iterations to learn the policy 

To improve exploration, we apply $\epsilon$-greedy strategy in the recognition
episode during data generation. 
The $\epsilon$-greedy strategy picks the action that has the maximum
$Q^{\pi}(s,a)$ with probability $1-\epsilon$ and a random action with probability
$\epsilon$. We use random policy for $\pi^{(0)}$ in the first iteration to
generate samples,
and we use all the samples generated during iteration $1 \sim i$ to
learn $\theta^{(i+1)}$.

\begin{figure}[t]
    \center
    \includegraphics[width=1.\linewidth]{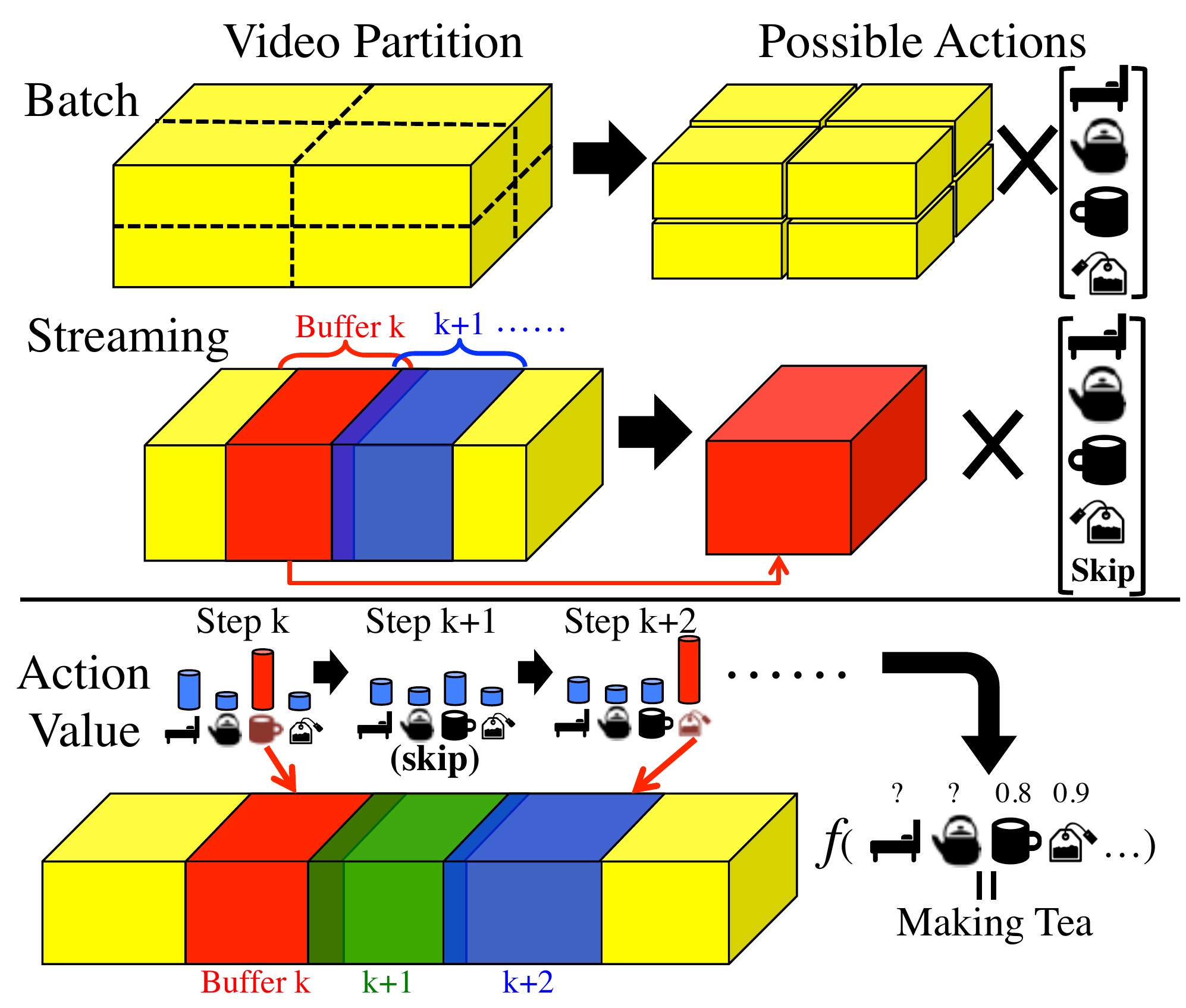}
    \caption{\label{fig:approach} Action spaces.
        Top: In batch, the whole video is divided into subvolumes, and actions
        are defined by the volume and object category to detect.  Middle: In
        streaming, the video is divided into segments by the buffer at each
        step, and actions are the object category to detect in the buffer plus
        a ``skip" action.  Bottom: Our method learns a video-specific policy to
        dynamically select a sequence of useful features to extract.
    }
\end{figure}

\subsubsection{Batch Recognition Setting}
\label{sec:batch}

In the \emph{batch recognition} setting, we have access to the entire test
video throughout an episode, and the budget is the total resources available
for feature computation, i.e., as capped by episode length $K$.  In this case,
our model is free to run an object detector at arbitrary locations.  Most
existing activity recognition work assumes this setting, though without
imposing a computation limit.  It captures the situation where one has an
archive of videos to be recognized offline, subject to real-world resource
constraints (e.g., auto-tagging YouTube clips under a budget of CPU time).

Each candidate location $l_m$ in the action set is a spatio-temporal volume.
Its position and size is specified relative to the length of the entire clip,
so that the number of possible actions is constant even though video lengths
may vary.  We use non-overlapping volumes splitting the video in half in each
dimension.  See Figure~\ref{fig:approach}, top.  Note that while the
bag-of-objects discards order, the action set \emph{preserves} it.  That means
our policy can learn to exploit the space-time layout of objects if/when
beneficial to feature prioritization (e.g., learning it is useful to look for a
washing machine \emph{after} a laundry basket, or an pot \emph{above} a stove).

In the batch setting, performing the same action at different steps in the
episode will produce the same observation.  Without loss of generality, we
define the time an action is performed as a constant $t^{k}{=}const. \, \forall
k$, and the action history feature $\delta t^{k}$ becomes a binary indicator
showing whether an action has been performed in the episode.  We forbid the
policy to choose actions that have been performed since they provide no new
information.

By design, the bag-of-objects is accumulated over time.  We impute the
observations of un-performed actions by exploiting previously learned object
co-occurrence statistics.  Let $\tilde{x} \in \mathbb{R}^{M}$ represent the
observation results of all actions on a video, where the $m$-th dimension
$\tilde{x}_{m} = x_{m}$ corresponds to the result of $m$-th action.  The vector
$\tilde{x}$ represents the object configuration in a video, and we learn its
probability $p(\tilde{x})$ on the same data that trains the activity recognizer
$f$ using a Gaussian Mixture Model (GMM):
\begin{equation}
    p(\tilde{x}) = \sum_{i=1}^{n} w_{i} \mathcal{N}(\tilde{x}|\mu_{i}, \Sigma_{i}),
\end{equation}
where we enforce a diagonal $\Sigma_{i}$ for computational efficiency.  At test
time, the model can be partitioned as
\begin{equation}
    \tilde{x} = \begin{bmatrix}
        \tilde{x}_{u} \\ \tilde{x}_{p}
    \end{bmatrix}, \;
    \mu_{i} = \begin{bmatrix}
        \mu_{iu} \\ \mu_{ip}
    \end{bmatrix}, \;
    \Sigma_{i} = \begin{pmatrix}
        \Sigma_{iu} & 0\\
        0 & \Sigma_{ip}
    \end{pmatrix},
\end{equation}
where $\tilde{x}_{p}$ corresponds to the observation results of performed
actions and $\tilde{x}_{u}$ to un-performed actions. 
We estimate $\tilde{x}_{u}$ using its expected value
over the conditional probability $p(\tilde{x}_{u}|\tilde{x}_{p})$, i.e.
\begin{equation}
    \langle \tilde{x}_{u} \rangle = \sum^{n}_{i=1} w^{\prime}_{i}\mu_{ip},
\end{equation}
where
\begin{equation}
    w^{\prime}_{i} =
    \frac{w_{i}\mathcal{N}(\tilde{x}_{p}|\mu_{ip},\Sigma_{ip})}
    {\sum_{i}w_{i}\mathcal{N}(\tilde{x}_{p}|\mu_{ip},\Sigma_{ip})}.
\end{equation}

\subsubsection{Streaming Recognition Setting}
\label{sec:streaming}

In the \emph{streaming} setting, recognition takes place at the same time the
video stream is received, so the model can only access frames received before
the current time step.  Further, the model has a fixed size buffer that
operates in a first-in-first-out manner; its feature requests may only refer to
frames in the current buffer.  Though largely unexplored for activity
recognition, the streaming scenario is critical for applications with stringent
resource constraints.  For example, when capturing long-term surveillance video
or wearable camera data, it may be necessary to make decisions online without
storing all the data.

The feature extractor can process a fixed number of frames per second, and this
rate indirectly determines the resource budget.  That is, the faster the
feature extractors can run, the more of them we can apply as the buffer moves
forward.  A recognition episode ends when it reaches the end of a video stream.

The action space consists of the $N$ object detectors (or alternatively, the
single CNN descriptor); an action's space-time location $l_m$ is always the
entire current buffer.  We further define a skip action $a_{0}$, which
instructs the model to wait until the next frame arrives without performing any
feature extraction.  Thus, for streaming, the number of actions equals the
number of objects plus one ($M{=}N{+}1$).  See Figure~\ref{fig:approach},
middle.  The skip action saves computation when the model expects a new
observation will not benefit the recognition task.  For example, if the model
is confident that the video is taken in a bedroom, and all un-observed objects
would appear only in the bathroom, then forcing the system to detect new
objects is wasteful.

Because new frames may arrive and old frames may be discarded during an action,
the video content available to the model will change between steps; performing
the same action at different steps yields different observations.  To connect
the video content in the buffer and the actions in the episode, we define the
time $t^{k}$ of the $k$-th action using the last frame number in the buffer
when the action was issued by the policy.

While we assume so far the video contains only the target activity, i.e.~the
video is trimmed to the span of the activity, our method generalizes to
\emph{untrimmed activity detection} in the streaming environment.  In that
case, the target activity only occurs in part of the video, and the system must
identify the span where the activity happens.  This is non-trivial in the
streaming environment.

To handle the streaming input, we pose the problem in terms of frame-level
labeling: we predict a label for each frame as it is received, and the activity
detector must optimize accuracy across all frames.  However, we do not estimate
the activity label from a single frame alone.  Rather, we predict each frame's
label using the temporal window around it.  For every newly arrived frame, we
consider all the windows shorter than an upper bound $\beta$ that end at the
frame. We predict the label of each window based on the same representation as
trimmed video, and we select the one with highest confidence as the prediction
result of the target frame.  Note that this requires storing only the
descriptors for recent history of length $\beta$, but keeping no video beyond
the current buffer.  The activity recognizer $f$ is a binary classifier trained
to determine whether the target activity occurs in the window, and actions are
terminated when a new frame arrives.

\section{Experiment}

\begin{figure*}[t]
    \center
    \includegraphics[width=1.\linewidth]{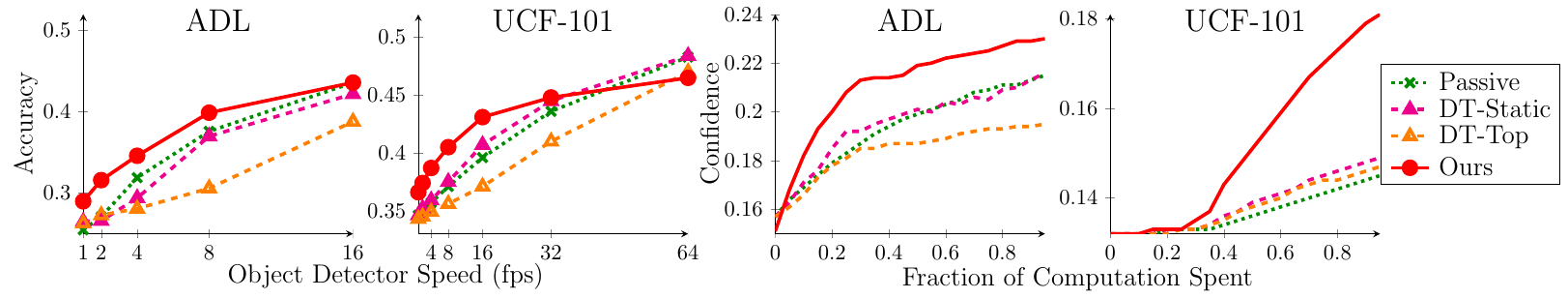}
    \caption{\label{fig:streams} Streaming recognition result. Left:
        Recognition accuracy as a function of object detector speed. Right:
        Confidence score improvement as the episode progresses.
    }
\end{figure*}

\subsection{Experiment Setting}
\paragraph{Datasets}

We evaluate on two datasets: the Activities of Daily
Living~\cite{pirsiavash2012cvpr} (ADL) and UCF-101~\cite{soomro2012arxiv}.
\textbf{ADL} consists of 313 egocentric videos recorded by 14 subjects, labeled
with $L{=}18$ activity categories (e.g., making coffee, using computer).
Following~\cite{pirsiavash2012cvpr}, we train $f$ in a leave-one-subject-out
manner.  Our policy is learned on a disjoint set of 110 clips (those used
in~\cite{pirsiavash2012cvpr} for training object detectors).  As observations
$x^{(k)}$, we use the provided object detector outputs for $N{=}26$ categories
(1 fps).  \textbf{UCF-101} consists of 13,320 YouTube videos covering $L{=}101$
activities.  We use the provided training splits to train $f$, reserve half of
the test splits for policy learning, and average results over all 6 splits.  As
observations $x^{(k)}$, we use the object detector outputs for $N{=}75$
objects, kindly shared by the authors of~\cite{jain2015cvpr}, which are
frame-level scores (no bbox).\footnote{We retain the 75 objects among all
    15,000 found most responsive for the activities,
following~\cite{jain2015cvpr}.  Because the provided detections are
frame-level, we split volumes only in the temporal dimension for $l_m$ on UCF.}
For CNN frame descriptors, we use the fc-7 activation of
VGG-16~\cite{simonyan2014arxiv} from Caffe Model Zoo (1 fps).  The video clips
average 78 and 19 seconds for ADL and UCF, respectively.

To create test data and policy learning data for the \emph{untrimmed}
experiments, we concatenate multiple clips
following~\cite{hoai2012cvpr,chaoyeh-cvpr2012}.  Although concatenation may
introduce discontinuity in content, it resembles scenarios in real videos.  For
example, it is similar to the video where the recorder walks from one room to
another and starts the next activity.  We concatenate five trimmed video clips
for one un-trimmed video. For each positive clip, we generate five un-trimmed
videos by placing the positive clip in different temporal location and drawing
four negative clips for other locations randomly.  We sort the categories by
their trimmed full observation results, and take the top 8 for untrimmed
experiments.  In all, we obtain 8,410 (UCF) and 3,130 (ADL)  untrimmed
sequences, with lengths averaging 2-7 minutes, respectively.  For all
experiments with trimmed data, (streaming/batch) we use the datasets as-is and
test all 18 (ADL) and 101 (UCF) activity categories.

\paragraph{Baselines}

We compare to several methods:
\begin{itemize}
\item \textbf{Passive}: selects the next action randomly.  It represents the
    most direct mapping of existing activity recognition methods to the
    resource-constrained regime.  The system does not actively decide which
    features to extract.

\item \textbf{Object-Preference}~\cite{jain2015cvpr}: a static feature
    selection heuristic employed for bag-of-objects activity recognition.  It
    prioritizes objects that appear frequently in each activity.  We average
    $x_m$ per activity and order $a_m$ based on its maximum response over all
    activities.  Though the authors intend this metric to identify the most
    discriminative objects---not to sequence feature extraction---it is a
    useful reference point for how far one can get with static feature
    selection.

\item \textbf{Decision tree (DT)}: a static feature ordering method.  We learn
    a DT to recognize activities, where the attribute space consists of the
    Cartesian product of object detectors and subvolume locations ($l_m$).  We
    sort the selected attributes by their Gini importance~\cite{breiman2001ml}.
    In the streaming case, we test two
    variations: DT-Static, where we cycle through the features in that order,
    and DT-Top, where we take only the top $P$ features and repeatedly apply
    all those object detectors on each frame.  $P$ is equal to the object
    detector framerate. Thus, DT-Top runs as many detectors as it can at
    framerate, prioritizing those expected to be most discriminative.

\item \textbf{Max-Margin Early Event Detector (MMED)}~\cite{hoai2012cvpr}: a
    state-of-the-art early event detector designed for untrimmed streaming
    video. It aims to fire on the activity as soon as possible after its onset.
    We implement it based on structure SVM solver
    BCFW~\cite{lacostejulien2013icml} and apply the authors' default parameter
    settings. The same window search process as in the untrimmed variant of our method is
    used for prediction, with a window size ranging from 1 to $\beta$ frames.

\end{itemize}

\paragraph{Implementation Details}

We run 8 iterations of policy iteration, with $\gamma{=}0.4$.  We initialize
$\epsilon{=}0.5$ for $\epsilon$-greedy exploration, and decrease by $0.1$ each
iteration with lower bound $0.05$.  For the streaming case, we use the video
framerate inherited from ADL (1 fps), and evaluate over a range of object
detector framerates.  We fix the buffer size to half the median clip length, 25
seconds.  We set the window size upper bound $\beta$ to one-third of the number
of object categories to avoid the model observing all objects within the
window.  For all methods, we initialize $\Psi(X)$ with features computed in the
first frame in the streaming case.

\subsection{Streaming Activity Recognition}
\label{sub:streaming_result}

First we test the streaming setting.  In this case, feature extraction speed
(e.g., object detector speed) dictates the action budget: the faster the
features can be extracted, the more can be used while keeping up with the
incoming video framerate.  We stress that to our knowledge, no prior activity
recognition work considers feature triage for streaming video.

Figure~\ref{fig:streams} (left 2 plots) shows the final recognition accuracy at
the episode's completion, as a function of the object detectors'
speed.\footnote{Object-Pref~\cite{jain2015cvpr} is not applicable to the
    streaming case because it lacks a unique object response prior for the
actions that is dependent on the buffer location.} Our method performs better
than the rest, across the range of detector speeds.  Overall, our method
reduces cost by 80\% and 50\% on UCF and ADL, respectively.  The left side of
the plots is most interesting; by definition all methods will converge in
accuracy once the object detector framerate equals the number of possible
objects to detect (26 for ADL and 75 for UCF).  DT-Top is the weakest method
for this task.  It repeatedly uses only the most informative features, but they
are insufficient to discriminate the 18 to 101 different activities.  This
result shows the necessity of instance-dependent feature selection, which our
method provides.

\begin{figure*}[t]
    \center
    \includegraphics[width=1.\linewidth]{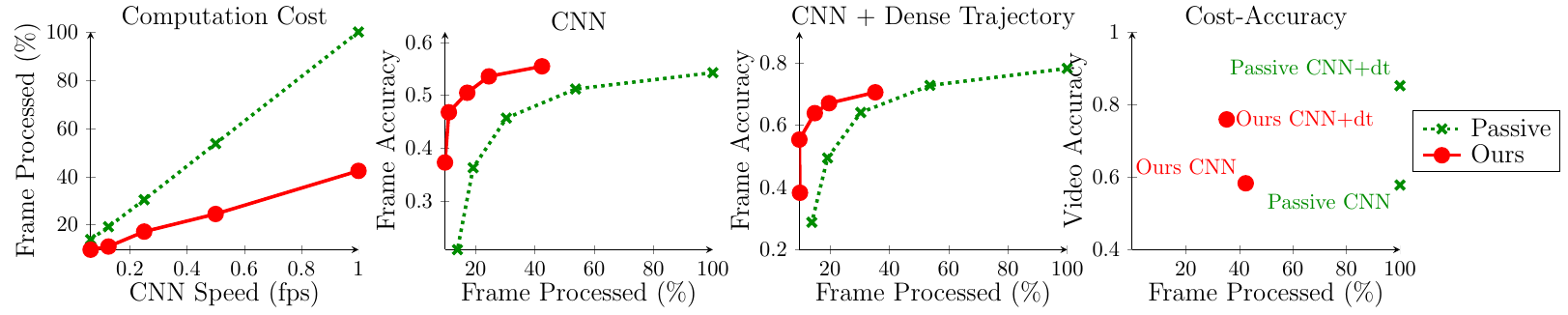}
    \caption{\label{fig:cnn_streams} Streaming recognition result on UCF-101
        using CNN frame features.
    }
\end{figure*}

Figure~\ref{fig:streams} (right 2 plots) shows the confidence score (of the
ground truth activity) improvement over the course of the episodes.  Here we
apply the 8 fps detector.  The baseline methods improve their prediction
smoothly, which indicates that they collect meaningful detection results at the
same rate throughout the episode.  In contrast, our method begins to improve
rapidly after some point in the episode. This shows that it starts to collect
more useful information once it has explored the novel video sufficiently.
Because UCF uses about 4$\times$ more objects in the representation, it takes
more computation (actions) before the representation converges.

\begin{table}[t]
    \begin{center}
    \begin{tabular}{lclll}
        \toprule
        $a^{(k)}$ & Result & Observed Obj. & Possible Activities & $a^{(k+1)}$\\
        \midrule
        \multirow{3}{*}{TV}  & + & None & Watch-TV & TV-remote\\
                             & - & Kettle & Watch-TV/Make-tea& Tea-bag\\
                             & - & Bottle & Drink-water & Fridge\\
        \midrule
        \multirow{3}{*}{Tap} & + & Dent-floss & Brush-teeth & Soap-passive \\
                             & - & Dish & Wash-dish/Watch-TV & Tap\\
                             & - & Soap-passive & Wash-hand & Soap-active\\
        \bottomrule
    \end{tabular}
    \caption{ \label{tab:learned_rules}Excerpts of policies learned from ADL in
 the streaming case. ``+'' and ``-'' indicate whether the object is detected at
 step-$k$.  Observed objects are those observed before $a^{(k)}$, and possible
 activities are the most likely activities predicted at step-$k$.
    }
    \end{center}
\end{table}

Table~\ref{tab:learned_rules} shows example excerpts of learned policies with
objects.  Here we see, for example, how our approach learns to detect objects
that can verify current activity hypothesis or differentiate ambiguous
activities, e.g., tap does not co-occur with TV, so seeing tap rules out
``Watch-TV.'' It also demonstrates detailed memory such that it looks for
objects that have been observed before but in a different status (actively
being used by the recorder vs.~passively sitting there).

\begin{figure*}[t]
    \center
    \includegraphics[width=1.0\linewidth]{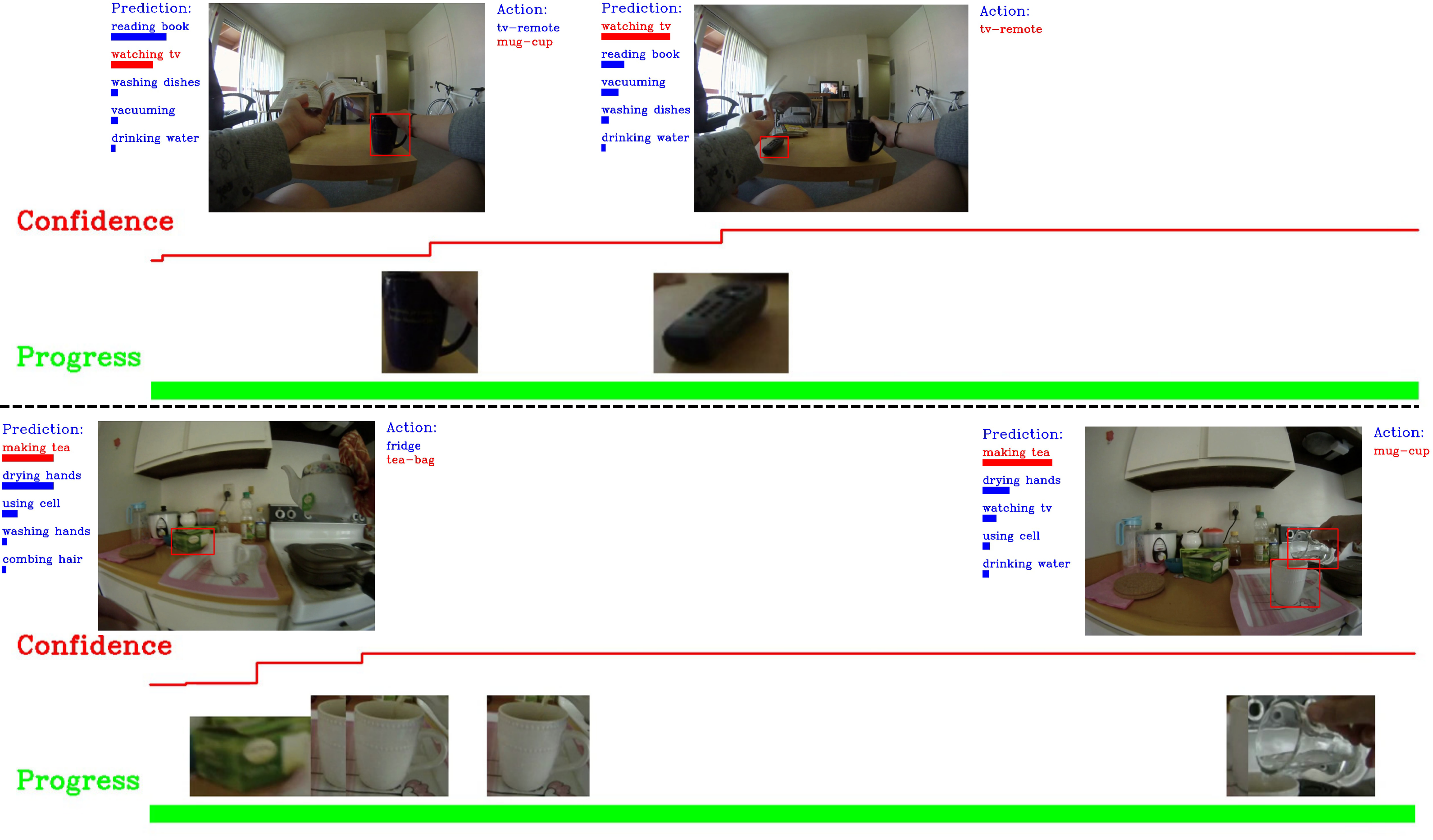}
    \caption{\label{fig:sample_adl_object} Two example recognition episodes on ADL
    with bag-of-object observation under streaming recognition setting.}
\end{figure*}

Next, we show the visual examples of the learned policy in action.
Figure~\ref{fig:sample_adl_object} shows the policy recognizing a video clip
from ADL with bag-of-object observation. In the first example, the policy
observes a mug-cup and identify the activity as either \textsl{reading book} or
\textsl{watching TV}. It then looks for tv-remote to disambiguate the two
activities. In the second example, the policy looks for tea-bag to recognize
whether the activity is \textsl{making tea} or \textsl{draying hand}. After
observing tea-bag, it looks for mug-cup to verify its prediction.

\begin{figure*}[t]
    \center
    \includegraphics[width=1.0\linewidth]{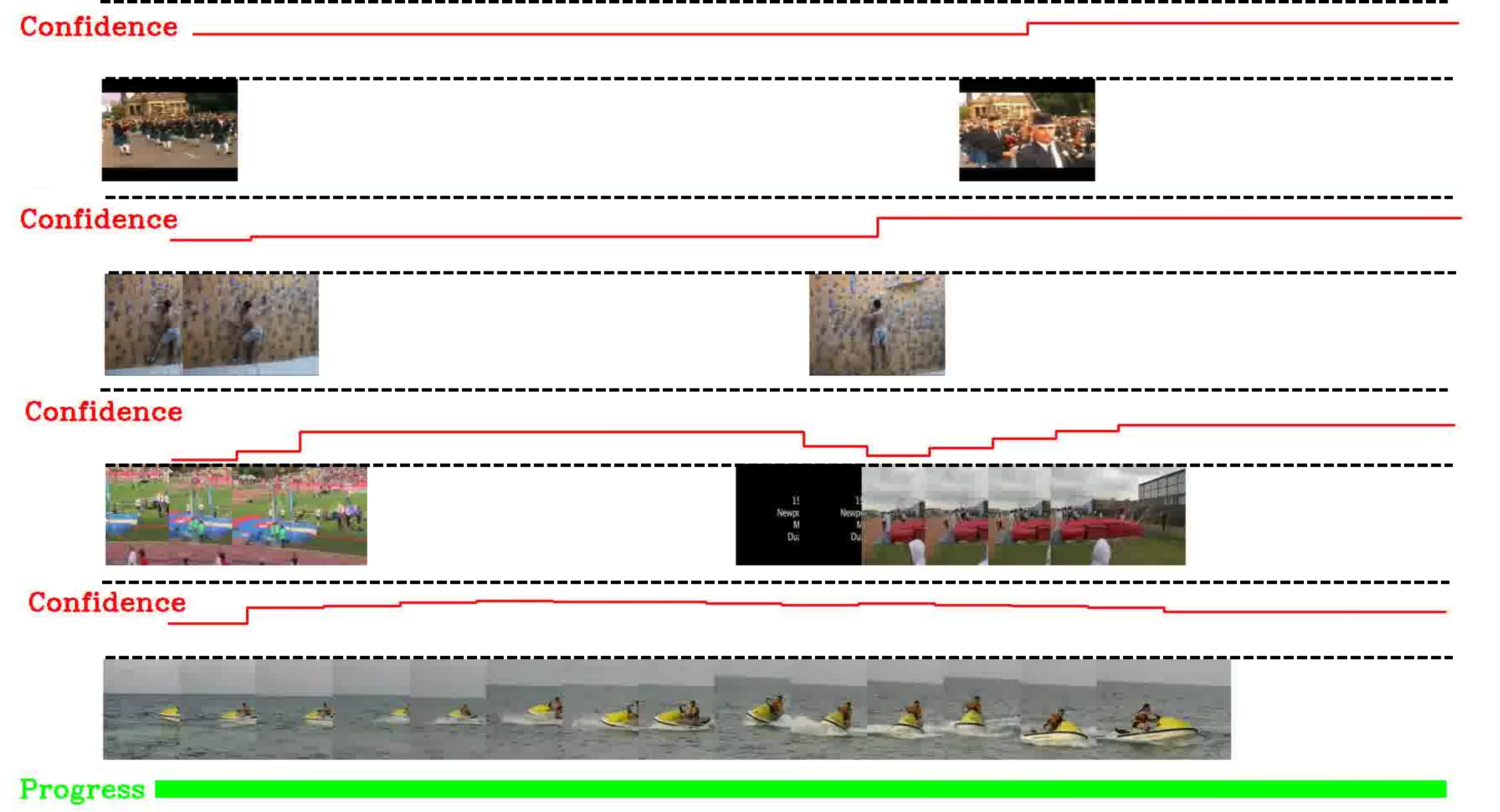}
    \caption{\label{fig:sample_ucf_cnn} Four example recognition episodes on UCF-101
        with CNN observation under streaming recognition setting.
    }
\end{figure*}

Figure~\ref{fig:sample_ucf_cnn} shows the policy recognizing video clips from
UCF-101 with per-frame CNN observation. The policy usually processes several
frames at the beginning and decides the following frames are unlikely to be
informative to the activity. Therefore, it starts skipping frames and resumes
processing at a more distant frames which may provide more distinct evidence
such as a closer view of the activity or different pose of the subject. From
the first three examples, we can see the number of frames skipped and the
number of frames processed is dependent on the observed content. For example,
in the third episode, the abrupt scene change decreases the  confidence
significantly, and the policy spends more computation to verify its prediction.
Finally, the last episode shows a failure case where the policy fails to stop
computation even if the prediction is fairly stable.

Figure~\ref{fig:cnn_streams} shows our method has clear advantages if applied
with CNN features as well.\footnote{ADL is less amenable to full-frame CNN
descriptors, due to domain shift of egocentric video and the nature of the
composite, object-driven activities.}  Here the DT baselines are not
applicable, since there is only one feature type; the question is whether to
extract it or not.  The Passive baseline uniformly distributes its frame
selections.  The left plot shows that no matter the framerate of the CNN
extractor, our method requires less than half of the frames to achieve the same
accuracy.  The second plot shows our method achieves peak accuracy looking
at just a fraction of the streaming frames, where the accuracy is measured over
every step in the recognition.  Our algorithm skips 80\% of the frames, but
still achieves over 90\% of the ultimate accuracy obtained using \emph{all}
frames.  With the base sampling rate of 1 fps, processing 20\% of the frames
means we extract features for only 0.8\% of the entire video. 

In the third plot, we further combine dense trajectories (dt) with the CNN
features to show that our method can benefit from more powerful features
without modification.  The right plot compares the cost-accuracy tradeoff
between the ultimate multi-class accuracy achieved by our streaming method
vs.~that attained using exhaustive feature extraction.  We obtain similar
accuracies with substantially less computation.

\begin{figure*}[t]
    \center
    \includegraphics[width=1.0\linewidth]{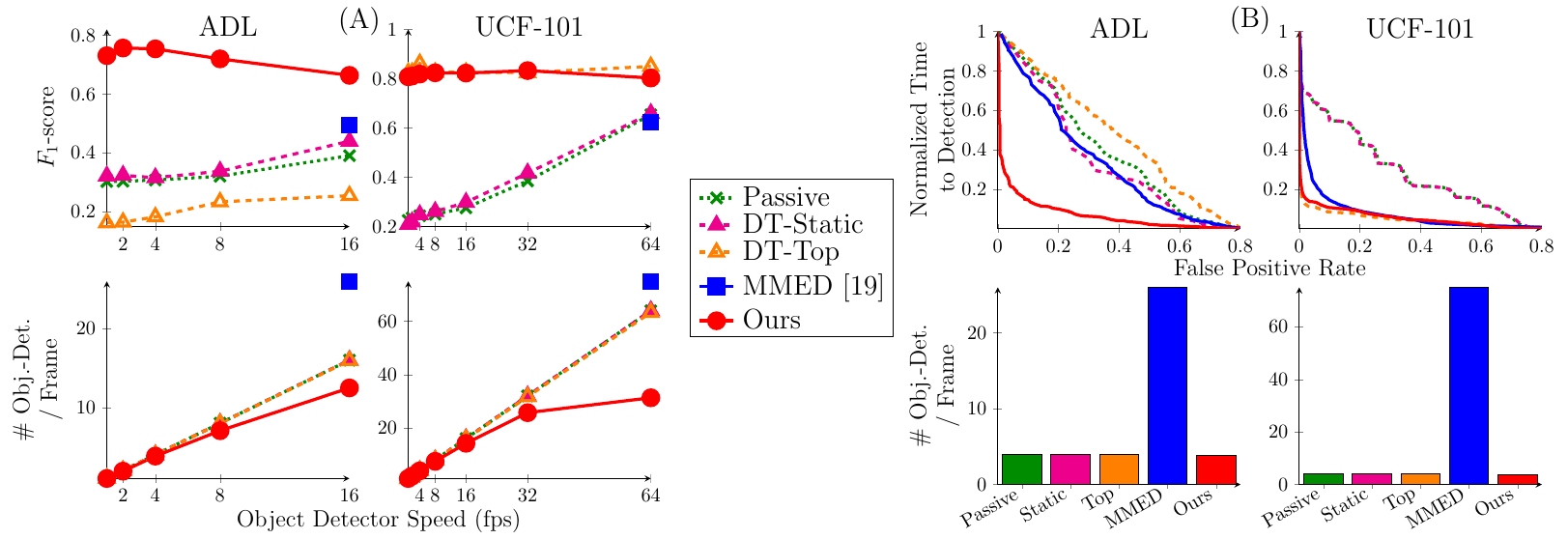}
    \caption{\label{fig:untrimmed}Streaming untrimmed detection results, with comparison to~\cite{hoai2012cvpr}. \textbf{(A)}:
        Accuracy (top, higher is better) and computation cost (bottom) as a
        function of object detector framerate.  \textbf{(B)}: Activity
        monitoring operating curves (top, lower is better) and corresponding
        computational costs (bottom) per method.
    }
\end{figure*}

\subsection{Untrimmed Video Activity Detection}

Next we evaluate streaming detection for untrimmed video.  This setting permits
comparison with the state of the art MMED~\cite{hoai2012cvpr} ``early''
activity detector.

Since we must predict whether each frame is encompassed by the target activity,
we measure accuracy with the $F_{1}$-score.  While we assume the episode
terminates after reaching the end of the video stream in our algorithm, in some
applications it may be sufficient to identify the occurrence of the activity
and then terminate the episode. Therefore, we further compare the detection
timeliness using the Activity Monitoring Operating Curve (AMOC),
following~\cite{hoai2012cvpr}.  AMOC is the normalized time to detection (NT2D)
vs.~the false positive rate curve.  The lower the value, the better the
timeliness of the detector.

In Figure~\ref{fig:untrimmed}(A), the top plots show the $F_{1}$-scores.
Overall, our method performs the best in terms of accuracy. On ADL, we achieve
nearly twice the accuracy of all baselines until the object detector speed
reaches 16 fps.  On UCF, our method is comparable to the best baseline, DT-Top.
Whereas DT-Top is weak on UCF for the multi-class recognition scenario (see
above), it fares well for binary detection on this dataset.  This is likely
because the UCF activities are often discriminated by one or few key objects,
and we give the baselines the advantage of pruning the object set to those most
responsive on each activity.

The bottom two plots in Figure~\ref{fig:untrimmed}(A) show the actual number of
object detectors run.  Our method reduces computation cost significantly under
high object detector speeds, thanks to its ability to forgo computation with
the ``skip" action.  In particular, it performs 50\% fewer detections under 64
fps on UCF while maintaining accuracy.  On the other hand, the baseline
methods' cost grows linearly with the object detector speed.

Figure~\ref{fig:untrimmed}(B) shows the AMOC under 4 fps detection speed (top,
see appendix for others) and the associated computational costs (bottom).
Despite the fact our reward function does not specifically target this metric,
our method achieves excellent timeliness in detection.  MMED performs second
best on the metric, but it incurs much higher computation cost than ours, as
shown by the bar charts.  This is because MMED is trained to fire early, but
always extracts all features in the frames it does process.

\subsection{Batch Activity Recognition}
\label{sub:batch_result} 

Finally, we test the batch setting.  We evaluate accuracy as a function of the
computation budget---the fraction of all possible actions the algorithm
performs (i.e., the number of features it extracts, normalized by video
length).  ``All possible" features would be extracting all features in all
frames (1 fps).  

\begin{figure*}[t]
    \center
    \includegraphics[width=1.\linewidth]{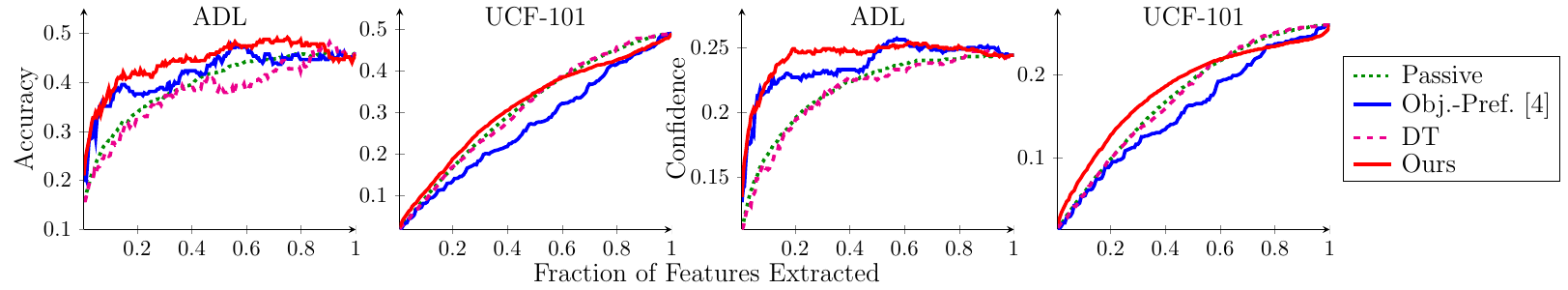}
    \caption{\label{fig:batch} Batch recognition accuracy/confidence score vs.~computational
        budget. 
    }
\end{figure*}

Figure~\ref{fig:batch} shows the results. Our method outperforms the baselines,
especially when the computation budget is low ($<0.5$).  In fact, extracting
only 30\% of the features on ADL, we achieve the same accuracy as with all
features.  Without a budget constraint, the video representation will converge
to that of the full observation---no matter what method is used; that is, all
methods must attain the same accuracy on the rightmost point on each plot.  Our
method shows more significant gains on ADL than UCF.  We think this reflects
the fact that the object categories for ADL are tailored well for the
activities (e.g., household items), whereas the object bank for UCF is more
diverse.   Furthermore, ADL has more objects in any single activity,  offering
more signal for our method to learn.  Object-Pref~\cite{jain2015cvpr}  is next
best on ADL, though it is noticeably weaker on UCF because it does not account
for the temporal redundancy of the dataset, i.e., a responsive object will be
equally responsive over the entire video. Our method is 2.5 times faster than
this nearest competing baseline.

Surprisingly, the Decision Tree (DT) baseline performs similarly to Passive.
(Note that DT-Static only is used; DT-Top is applicable only for the streaming
case.)  We attempted to improve its accuracy by learning it on the same
features as $f$, i.e., dropping the subvolumes from the attributes and running
one object detector over the entire video for each action.  However, this
turned out to be worse due to redundant/wasteful detections.  This shows the
importance of coping with partially observed results, which the proposed method
can do.

Our contribution is not a new model for activity recognition, but instead a
method that enables activity recognition for existing features/classifiers
without exhaustive feature computation.  This means the accuracies achieved
with ``all features" is the key yardstick to hold our results against.
Nonetheless, to put in context with other systems: the base batch recognition
model we employ gets results slightly better than the state-of-the-art on
ADL~\cite{pirsiavash2012cvpr,mccandless} and within 4.5-11\% of the
state-of-the-art using comparable features on
UCF~\cite{jain2015cvpr,simonyan-activity}(see Figure~\ref{fig:cnn_streams},
right two plots).  We suspect the UCF gap is due to our use of max-pooling
(vs.~average) and logistic regression.

\section{Conclusions}
We developed a dynamic feature extraction strategy for activity recognition
under computational constraints.  On two diverse datasets, our method shows
competitive recognition performance under various resource limitations.  It can
be used to consistently achieve \emph{better accuracy} under the same resource
constraint, or meet a given accuracy using \emph{less resources}.  In future
work we plan to investigate policies that reason about variable cost
descriptors.

\section*{Acknowledgements}
This research is supported in part by ONR PECASE N00014-15-1-2291 and a gift
from Intel.

%
\IEEEpeerreviewmaketitle

\appendices

\section{Streaming Recognition}
\label{sec:streaming_supp}

\begin{figure*}[t]
    \center
    \includegraphics[width=1.0\linewidth]{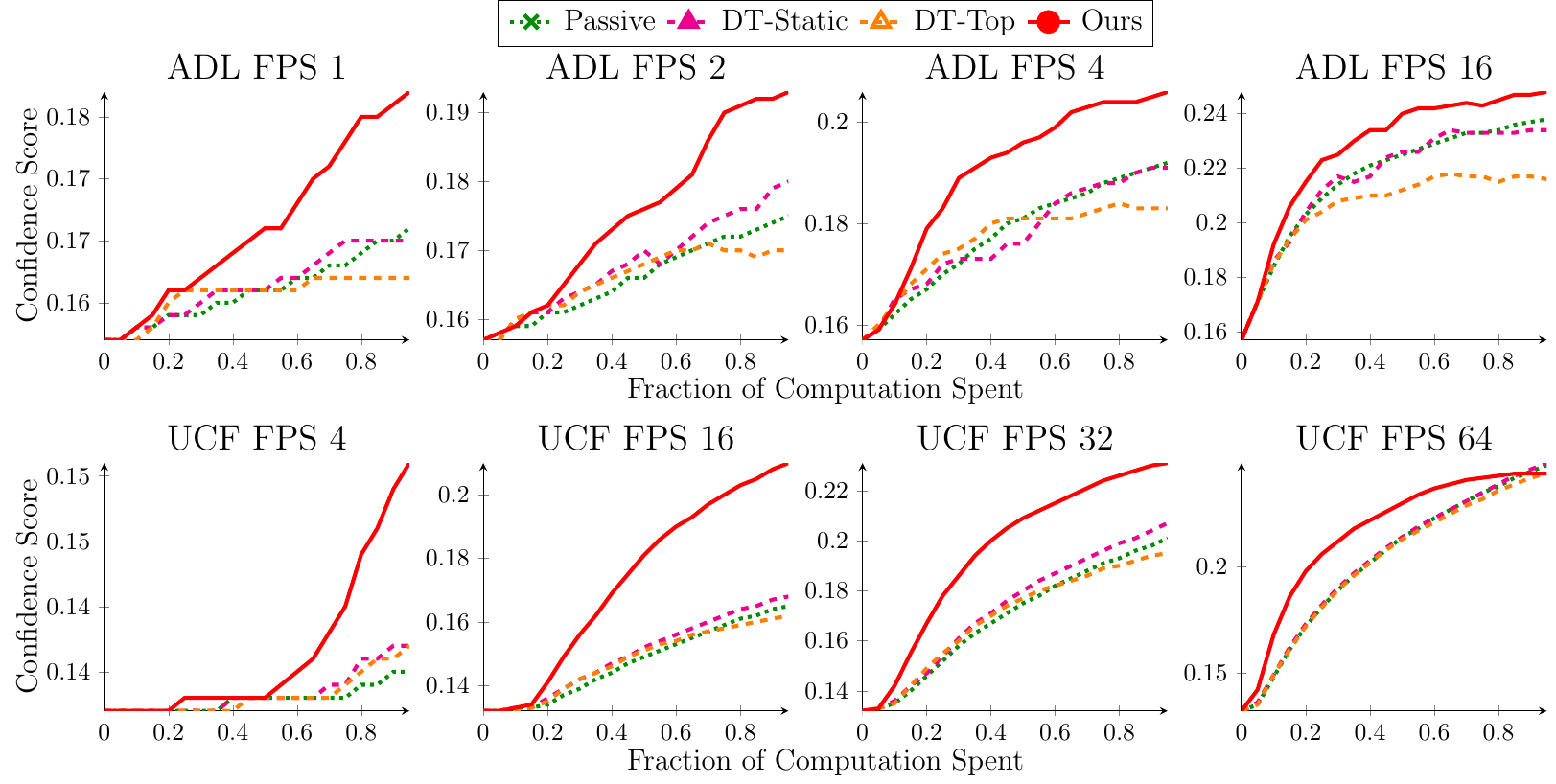}
    \caption{\label{fig:streaming_supp} Streaming recognition accuracy under
        different object detector speed. These plots go with the one in
        Figure~\ref{fig:streams} above.
    }
\end{figure*}

We show the confidence score improvement during recognition episodes with an 8 fps
object detector speed in Section~\ref{sub:streaming_result}. For other object
detector speeds, please refer to Figure~\ref{fig:streaming_supp}. The results
are consistent with that of 8 fps, where our method performs better than others
under all object detector speeds, and the performance of different methods
become more similar as the detector speed becomes faster. We do not show the
results of 1 and 2 fps on UCF, because UCF videos are on average shorter, and
for detectors that slow the recognition episodes consist of single action for
videos shorter than the buffer size, making the curves meaningless.

Note the number of object $N{=}26$ for ADL and $N{=}75$ for UCF, and using
object detector speed that exceed the number of object will reduce the problem
to full observation of the video. Therefore, we show 32 fps and 64 fps results
only for UCF.

\begin{figure*}[t]
    \center
    \includegraphics[width=1.0\linewidth]{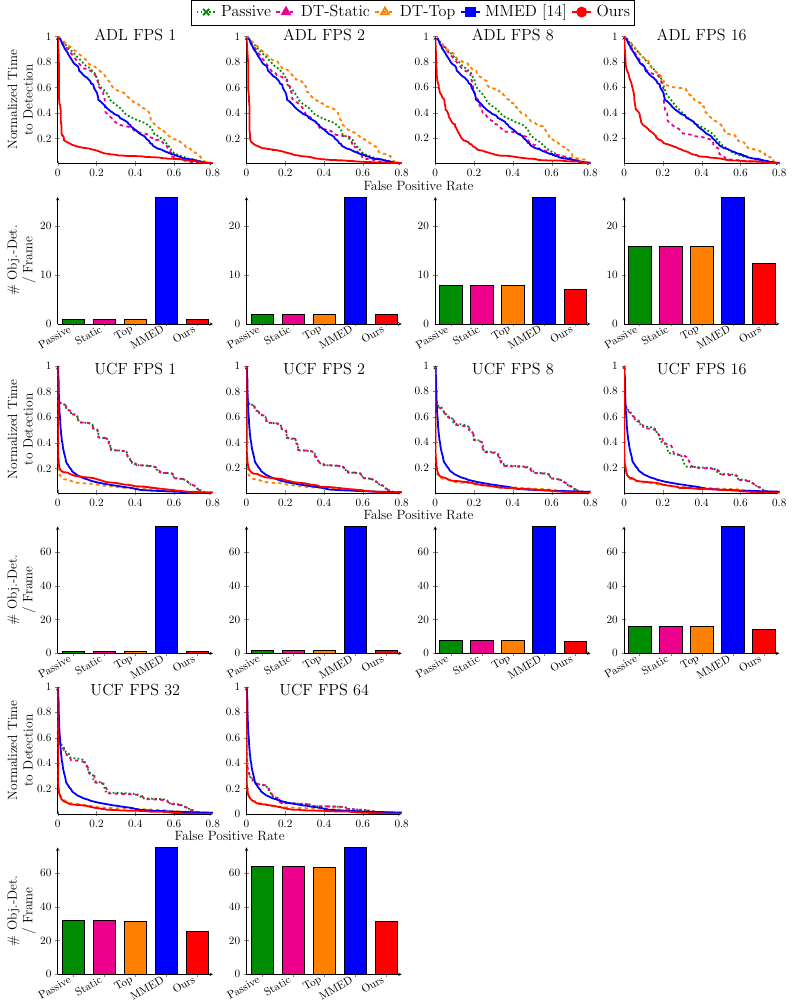}
    \caption{\label{fig:amoc} AMOC under different object detector speed. These
        plots go with the ones in Figure~\ref{fig:untrimmed} above.
    }
\end{figure*}

\section{Un-trimmed Video Activity Detection}
\label{sec:untrimmed_supp}

In Figure~\ref{fig:untrimmed}, we show AMOC under 4 fps object detector speed.
For the complete result, please refer to Figure~\ref{fig:amoc} which shows AMOC
under all other object detector speeds.  Similar to the result in the paper,
our method achieves excellent timeliness under all object detector speeds.
Also, we can see more clearly how our method reduces computational cost under a
high object detector speed. It uses only half of the computation on UCF under
64 fps object detector speed while remaining the best performing method.

\ifCLASSOPTIONcaptionsoff
  \newpage
\fi



%

\bibliographystyle{IEEEtran}
\bibliography{videorecognition}

\end{document}